\documentclass{edm_article}
\usepackage{colortbl}
\usepackage{xcolor}
\usepackage{graphicx}
\usepackage{hyperref}
\usepackage{hhline}
\usepackage{multirow}
\usepackage{blindtext}
\usepackage{multicol}
\usepackage{diagbox}
\usepackage{booktabs}
\usepackage{breakurl}
\usepackage{url}
\usepackage{threeparttable}
\newcolumntype{M}[1]{>{\centering\arraybackslash}m{#1}}

\definecolor{effort}{HTML}{6DFCFF}
\definecolor{outcome}{HTML}{F4CCCC}

\definecolor{royal_purple}{RGB}{0, 0, 0}

\begin{document}



\title{How Can I Improve? Using GPT to Highlight the Desired and Undesired Parts of Open-ended Responses}

\numberofauthors{8}
\author{
\alignauthor Jionghao Lin\\
       \affaddr{Carnegie Mellon University}\\
       \email{jionghal@cs.cmu.edu}
\and
\alignauthor Eason Chen\\
       \affaddr{Carnegie Mellon University}\\
       \email{easonc13@cmu.edu}
\and
\alignauthor Zeifei Han\\
       \affaddr{University of Toronto}\\
       \email{feifei.han@mail.utoronto.ca}
\and
\alignauthor Ashish Gurung\\
       \affaddr{Carnegie Mellon University}\\
       \email{agurung@andrew.cmu.edu}
\and
\alignauthor Danielle R. Thomas\\
       \affaddr{Carnegie Mellon University}\\
       \email{drthomas@cmu.edu}     
\and
\alignauthor Wei Tan\\
       \affaddr{Monash University}\\
       \email{wei.tan2@monash.edu}
\and
\alignauthor Ngoc Dang Nguyen\\
       \affaddr{Monash University}\\
       \email{dan.nguyen2@monash.edu}
\and
\alignauthor Kenneth R. Koedinger\\
       \affaddr{Carnegie Mellon University}\\
       \email{koedinger@cmu.edu}
}

\maketitle

\begin{abstract}

Automated explanatory feedback systems play a crucial role in facilitating learning for a large cohort of learners by offering feedback that incorporates explanations, significantly enhancing the learning process. However, delivering such explanatory feedback in real-time poses challenges, particularly when high classification accuracy for domain-specific, nuanced responses is essential. Our study leverages the capabilities of large language models, specifically Generative Pre-Trained Transformers (GPT), to explore a sequence labeling approach focused on identifying components of desired and less desired praise for providing explanatory feedback within a tutor training dataset. Our aim is to equip tutors with actionable, explanatory feedback during online training lessons. To investigate the potential of GPT models for providing the explanatory feedback, we employed two commonly-used approaches: \textit{prompting} and \textit{fine-tuning}. To quantify the quality of highlighted praise components identified by GPT models, we introduced a Modified Intersection over Union (M-IoU) score. Our findings demonstrate that: (1) the M-IoU score effectively correlates with human judgment in evaluating sequence quality; (2) using two-shot prompting on GPT-3.5 resulted in decent performance in recognizing effort-based (M-IoU of 0.46) and outcome-based praise (M-IoU of 0.68); and (3) our optimally fine-tuned GPT-3.5 model achieved M-IoU scores of 0.64 for effort-based praise and 0.84 for outcome-based praise, aligning with the satisfaction levels evaluated by human coders. Our results show promise for using GPT models to provide feedback that focuses on specific elements in their open-ended responses that are desirable or could use improvement.




\end{abstract}


\section{Introduction}
Tutoring is an important instructional method that can be highly effective in supporting students. Tutors utilize various tutoring strategies to effectively facilitate learning opportunities~\cite{kraft2021blueprint, nickow2020impressive, lin2022good}. While the effectiveness of tutoring is widely recognized, various logistical challenges have restricted its widespread implementation. Specifically, recruiting, training, and retention of tutors have presented major hurdles~\cite{thomas2023tutor}. Training tutors can be highly resource-intensive and often requires hands-on training from experienced tutors. A key component of effective tutor training involves helping novice tutors learn effective tutoring strategies~\cite{lin2023using, thomas2023tutor}. For instance, instead of simply acknowledging an incorrect answer, effective tutors often engage with the student to identify the underlying misconceptions or gaps in knowledge that can provide additional context to the incorrect answer. This contextual insight can then assist the tutor in providing more effective support. Traditionally, these types of nuanced insights have been facilitated through hands-on training from more experienced tutors. However, the scalability of this hands-on approach remains a well-recognized limitation~\cite{lin2023using, hirunyasiri2023comparative, kakarla2024using, lin2024ijaied}, necessitating innovative solutions to extend this model of training tutors without compromising the quality of feedback.

\textcolor{royal_purple}{In response to the growing need for scalable hands-on support in tutor training, researchers have increasingly turned to automated feedback systems. The integration of such systems to enhance feedback is well-established within Educational Data Mining (EDM), with numerous studies demonstrating their efficacy \cite{lin2023learner, beck2008does, gurung2023common, patikorn2020effectiveness}. While many implementations have employed AI algorithms to generate automated feedback \cite{cavalcanti2021automatic}, the specific application to tutor training remains underexplored. In this emerging field, the development of automated explanatory feedback systems designed for tutors presents a promising avenue. An illustrative example includes work by \cite{lin2023using}, which utilized the BERT language model \cite{devlin2019bert} to enhance tutor training. Although the results showed potential, a significant challenge emerged: The BERT model was hampered by a lack of access to extensive datasets, limiting its ability to offer precise, context-specific feedback. This challenge is similarly problematic for other traditional models such as Conditional Random Fields (CRF) and Hidden Markov Models (HMM), which also require adequate domain-specific training data \cite{nguyen2023low, luo2023re, Nguyen_Tan_Du_Buntine_Beare_Chen_2023}. Recent advances in large language models (LLMs) present a viable solution to these challenges. LLMs, such as Generative Pre-trained Transformers (GPT) developed by OpenAI, are pre-trained on vast and diverse datasets, enabling them to generalize effectively across different domains without extensive task-specific data. The inherent adaptability of GPT models to dynamically adjust to specific contextual scenarios makes them well-suited for developing real-time, tailored feedback systems for tutor training—offering the adaptive, hands-on support that models like BERT could not.}




By referencing recent LLM literature \cite{jurafsky2022speech, pornprasit2024gpt, kalyan2023survey}, we explore two approaches to leverage the potentials of GPT models in educational contexts: \textit{prompting} and \textit{fine-tuning}. Prompting involves designing input queries that guide the GPT model to generate desired outputs by leveraging its pre-existing knowledge and capabilities \cite{jurafsky2022speech, kalyan2023survey}. This approach is particularly useful for tasks requiring immediate, context-specific responses without the need for extensive model retraining. In comparison, fine-tuning adjusts the model's parameters on a targeted dataset, thereby optimizing its performance for specific tasks or domains \cite{jurafsky2022speech, kalyan2023survey}. The fine-tuning approach allows for a more tailored response generation, closely aligned with the nuances of the given context. Both approaches exhibit significant promise in text comprehension and generation, suggesting their potential effectiveness in producing nuanced, explanatory feedback. Thus, our study aims to harness these approaches to unveil the full capacity of GPT models in automating the generation of high-quality explanatory feedback, thereby addressing a critical need in educational feedback systems. Driven by this, our study proposed two \textbf{R}esearch \textbf{Q}uestions (\textbf{RQ}s):

\textbf{RQ1:} To what extent can we prompt the GPT models to enhance the prediction accuracy of providing explanatory feedback?\\
\textbf{RQ2:} To what extent can the fine-tuned GPT models enhance the prediction accuracy of providing explanatory feedback?

\begin{figure}[h]
\centering
\includegraphics[width=0.46\textwidth]{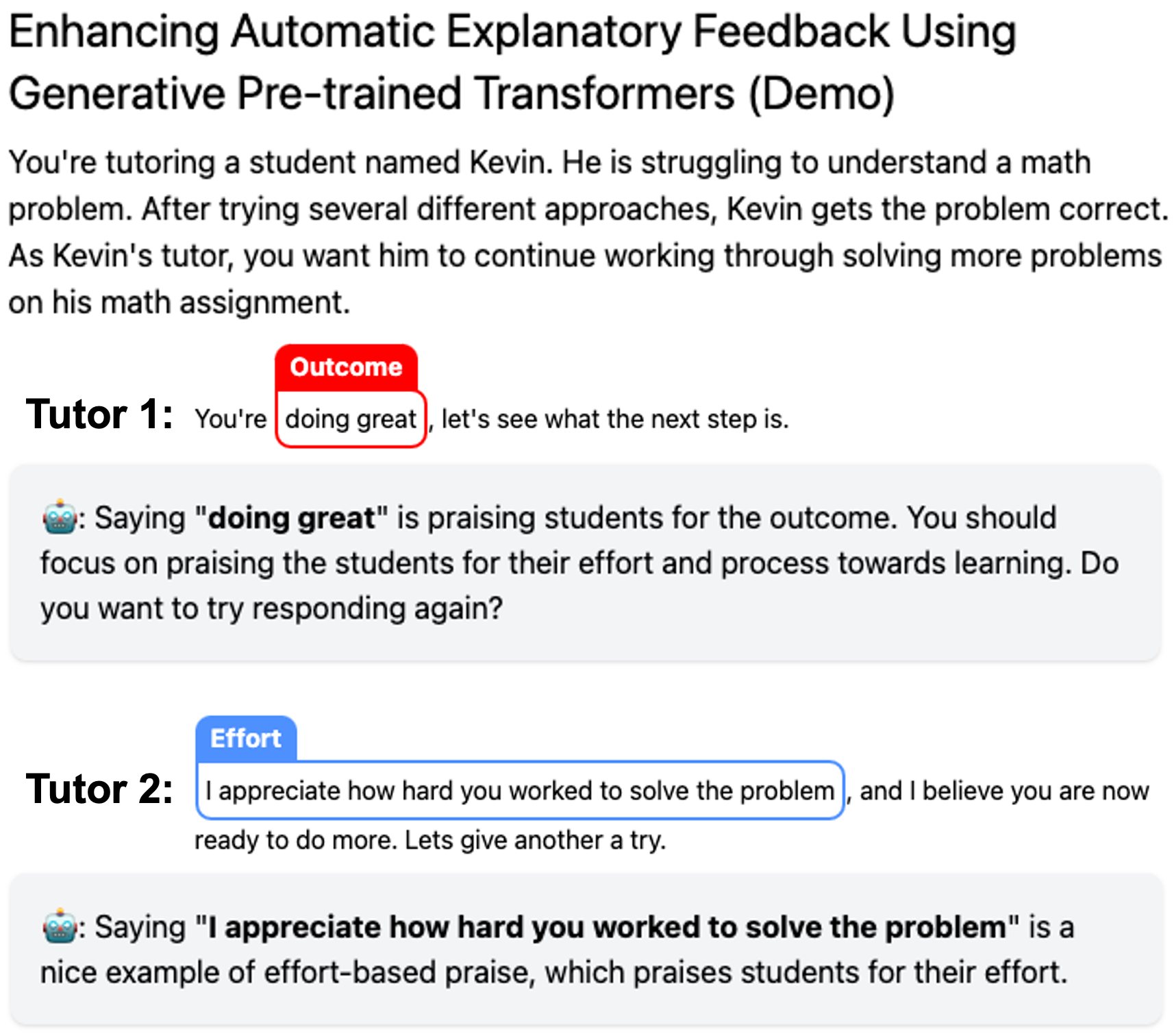}
\caption{Automated explanatory feedback using sequence labeling method facilitated by fine-tuned GPT-3.5 model} \label{demo_pic}
\end{figure}

Through this work, we aim to offer a scalable solution that enhances tutor training programs and, ultimately, the learning experience for students. Our study developed an automated explanatory feedback system to highlight the correct and incorrect components of praise from novice tutor attempts, as illustrated in Figure \ref{demo_pic}. We implemented sequence labeling method for highlighting the correct and incorrect components by using the approaches of prompting and fine-tuning GPT models. To evaluate the quality of highlighted praise components from tutor responses by GPT models, we introduced the Modified Intersection over Union (M-IoU) score, a metric designed for our task. Our results indicate a strong correlation between the M-IoU score and human evaluators' judgments regarding the quality of highlights, affirming the metric's reliability.

In addressing \textbf{RQ1}, we employed a two-shot prompting method to prompt GPT-3.5 and GPT-4 models to highlight desired and undesired components of praise in tutor responses. Notably, the GPT-3.5 model demonstrated performance on par with that of the GPT-4 model, exhibiting commendable accuracy in identifying effort-based (M-IoU of 0.46) and outcome-based praise (M-IoU of 0.68). These levels of accuracy are considered decent by human coders, highlighting the effectiveness of our prompting strategies. For \textbf{RQ2}, we delved into fine-tuning the GPT models across a set of training sample sizes—from 10\% of our dataset (13 samples) to 50\% (65 samples)—to gauge how fine-tuning influences the model's ability to enhance the precision of explanatory feedback. Due to limitations in accessing the fine-tuning GPT-4 model, our investigation focused on fine-tuning the GPT-3.5 model. The optimal fine-tuned GPT-3.5 model achieved M-IoU scores of 0.64 for effort-based praise and 0.84 for outcome-based praise, aligning with the satisfaction levels observed by human coders. Motivated by the effectiveness of our fine-tuned GPT model, we have built a demo\footnote{You can experience the promising features and limitations of our existing system in this demo \url{https://edm24-effort-outcome.vercel.app/}} of our automated explanatory feedback system.


\section{Background}
\subsection{Effective Tutoring Practice}
Effective tutoring plays an important role in enhancing student learning by integrating academic knowledge with the capability to address students' socio-motivational needs \cite{dietrichson2017academic, guryan2023not, nickow2020impressive, lin2022good}.  However, equipping tutors with these skills proves challenging, given the limited active learning opportunities that bring situational, scenario-based experiences to the professional development of tutors \cite{chine2022development}. Thus, current tutor training for tutors in addressing the social-emotional and motivational aspects of student learning remain underdeveloped\cite{chine2022development, reich2022teaching}. 

Our study focuses on a particular aspect of tutoring practice: the delivery of effective praise. Praise is a fundamental tutoring practice during the human tutoring process, consistently shown to have a positive impact on student motivation, engagement, and learning outcomes\cite{jenkins2015rates, kamins1999person, thomas2023tutor}. Research highlights that for praise to be effective, it should be: (1) sincere, earned, and truthful; (2) specific by giving details of a student's strengths; (3) immediate, with praise given right after the student's action; (4) authentic, avoiding repetitive phrases like \textit{``great job''} which diminishes meaning and becomes predictable, and (5) focused on the learning process rather than innate ability \cite{thomas2023tutor}. Existing literature categorizes praise into three types: effort-based \textit{Effort}, outcome-based \textit{Outcome}, and person-based \textit{Person} \cite{kamins1999person, thomas2023tutor, chine2022scenario, chine2022development}. Effort-based praise emphasizes the student's learning process (e.g., \textit{``I love your effort that you put into the writing...''}). Outcome-based praise highlights a student's achievements, like scoring high on an assignment or solving a problem correctly, and it's sometimes linked to less effective praise strategies such as \textit{``Good job!''}. Person-based praise attributes success to innate qualities beyond the student's control such as \textit{``You are smart!''} and is often, similar to outcome-focused praise, considered less effective \cite{kamins1999person}.

Training novice tutors to provide more effective praise (i.e., effort-based praise) requires a comprehensive understanding of both the desirable and less favorable elements of their praise responses. For tutors to refine their skills effectively, they should engage in a feedback process to know how well their responses align with the effective praise in tutoring\cite{chine2022development, thomas2023tutor}. However, manual feedback generation by expert tutors poses significant challenges due to its time-consuming and labor-intensive nature. This underscores the necessity for exploring automated feedback systems within tutor training programs. Such systems could offer scalable and timely feedback, thereby enhancing tutors' ability to effectively address student motivation issues.



\subsection{Feedback for Tutor Training}
Feedback in the learning process is universally recognized for its significant impact on learning outcomes \cite{patikorn2020effectiveness, gurung2023common, gurung2023identification, lin2023learner, henderson2019impact, hattie2007power}, with effects ranging from significantly positive \cite{patikorn2020effectiveness, gurung2023common} to occasionally negative \cite{gurung2023identification}, depending on the content and method of delivery. The effectiveness of feedback, as highlighted by Hattie and Timperley \cite{hattie2007power}, is intricately linked to its relevance to the learning context, its timing following initial instruction, and its focus on addressing misconceptions or incorrect reasoning \cite{hattie2007power}. In particular, immediate, explanatory feedback, which clarifies why certain responses are correct or incorrect, plays a crucial role in promoting active engagement and thoughtful practice among learners \cite{ryan2021designing, lin2023learner, hattie2007power, henderson2019impact, gurung2023common, patikorn2020effectiveness}. The growing importance of feedback has motivated the adoption of automated feedback systems in educational settings, such as OnTask, which allows educators to provide scalable feedback based on conditional rules related to students' academic activities and performance \cite{pardo2018ontask}. Yet, the application of such systems in tutor training remains under-explored.

An important method of deploying automated feedback in tutor training involves the use of templated feedback. The templated feedback, including specific references to desired and less-desired elements of the tutor responses, is informed by earlier results on the effectiveness of having a rich, data-driven error diagnosis taxonomy driving template-based feedback \cite{aleven2001towards}. Our study aims to employ natural language processing (NLP) techniques to automate the identification of desirable and less desirable elements within tutor responses, facilitating the generation of templated explanatory feedback.


\subsection{Sequence Labeling for Feedback Generation}
Sequence labeling, a fundamental task in natural language processing (NLP), plays a pivotal role in identifying and categorizing key segments of text according to predefined labels \cite{jurafsky2022speech}. To elucidate the mechanism of sequence labeling, we consider Named Entity Recognition (NER) as a representative subtask, which is closed to the task in our study. NER seeks to automatically detect and classify named entities—words or phrases with specific attributes—into categories such as person, organization, and location \cite{jurafsky2022speech, li2020survey}. For instance, in the sentence \textit{``John said that Pittsburgh is wonderful in the winter,''} the terms \textit{``John''}, \textit{``Pittsburgh''}, and \textit{``winter''} would be labeled as \texttt{Person}, \texttt{Location}, and \texttt{Time}, respectively, showcasing how NER distinguishes and categorizes entities within a textual context.

Our study extends the application of sequence labeling to identify and highlight components related to different types of praise within tutor responses. This process involves discerning specific words or phrases that signify the kind of praise being used, thereby offering tutors insight into their feedback practices. For example, \textit{``You are doing great.''}, the phrase \textit{``doing great''} in this context is identified as an outcome-based praise. Leveraging sequence labeling allows our AI model to spotlight such instances of praise, enabling the provision of nuanced, explanatory feedback. An example of such feedback might be,``\textit{Saying ``doing great'' is praising the student for the outcome. You should focus on praising the students for their effort and process towards learning. Do you want to try responding again?}'' This approach facilitates the generation of targeted, template-based feedback for tutors. 

Notably, while previous research has explored sequence labeling techniques for similar purposes \cite{lin2023using}, the accuracy of their proposed models in precisely identifying and categorizing feedback elements remains a challenge. This limitation underscores the need for leveraging more advanced models to provide accurate, informative feedback to tutors.

\subsection{Large Language Models in Education}
Recent advancements in natural language processing have seen the evaluation of large language models (LLMs) like GPT models in various educational tasks, leveraging techniques such as prompting and fine-tuning \cite{kasneci2023chatgpt}. The GPT models (e.g., GPT-3.5 or GPT-4) have demonstrated significant potential in enhancing many educational tasks (e.g., feedback generation and learning content generation) \cite{kasneci2023chatgpt}. Motivated by these developments, our study aims to investigate the applicability of prompting and fine-tuning GPT models to identify both desirable and less desirable aspects of tutoring responses. We intend to evaluate the effectiveness of these approaches in developing an automated system for providing explanatory feedback.

 
\subsubsection{Prompting large language models}

Prompting, which involves the use of specific queries or statements to guide a large language model’s (LLM) output, has been identified as a significant technique for leveraging the capabilities of LLMs in education \cite{kasneci2023chatgpt}. The prompting strategy plays a pivotal role in effectively guiding the models, such as GPT-3 and GPT-4, to produce responses that are more aligned with the context and requirements of the tasks. Research by Dai \textit{et al.} \cite{daiaassessing} on the GPT-3.5 and GPT-4 model highlighted GPT models' ability to generate student feedback that surpassed human instructors in readability. Furthermore, Hirunyasiri \textit{et al.} \cite{hirunyasiri2023comparative} demonstrated the superiority of the GPT-4 model over human expert tutors in assessing specific tutoring practices. \cite{levonian2023retrievalaugmented} used GPT-4 model to generate high-quality answer responses for middle school math questions. \cite{mcnichols2023exploring} providing feedback for multiple-choice questions at the middle-school math level. Given that GPT models has shown remarkable performance on various educational tasks \cite{dai2023can, hirunyasiri2023comparative, levonian2023retrievalaugmented, mcnichols2023exploring}, our study also leveraged the GPT models to further explore its capabilities in automatically generating explanatory feedback since the exploration of prompting GPT models for providing explanatory feedback in response to open-ended questions remains limited.



\subsubsection{Fine-tuning large language models}

In addition to prompting the GPT models, the fine-tuning of GPT models has also shown considerable promise in various educational tasks \cite{kasneci2023chatgpt}. The fine-tuning method adjusts the model's neural network to better suit particular domain, thereby enhancing its performance in relevant contexts \cite{jurafsky2022speech}. Latif and Zhai \cite{latif2024fine} employed fine-tuned GPT-3.5 model for the purpose of automatic scoring in science education. Their findings indicate that GPT-3.5, once fine-tuned with domain-specific data, not only surpassed the performance of the established BERT model \cite{devlin2019bert} but also demonstrated superior accuracy in assessing a variety of science education tasks. Such advancements underscore the value of fine-tuning GPT models for educational applications, showcasing their ability to provide precise, scalable solutions across diverse educational settings. Bhat \textit{et al.} \cite{bhat2022towards} introduced a method for generating assessment questions from text-based learning materials using a fine-tuned GPT-3 model. The generated questions was further assessed regard to their usefulness to the learning outcome by human experts, with the findings revealing a favorable reception among human experts. Inspired by these pioneering research, our study aims to extend the application of fine-tuning method to GPT models within the context of generating explanatory feedback. While the aforementioned studies \cite{devlin2019bert, latif2024fine} have not directly addressed the generation of explanatory feedback, their success in applying fine-tuned LLMs within educational domains suggests a promising avenue for our investigation. By customizing GPT models to the nuances of educational feedback, we anticipate uncovering new potentials in automating and enhancing the feedback process. These efforts will contribute to the growing body of evidence supporting the integration of fine-tuned LLMs in educational technology, potentially revolutionizing the way feedback is generated and applied in learning environments.

\section{Method}
\subsection{Dataset}
\label{data}
\textcolor{royal_purple}{Our study received IRB ethical approval with the protocol number: STUDY2018\_00000287 from Carnegie Mellon University.} The study utilized a dataset comprising responses from 65 volunteer tutors who participated in the \textit{Giving Effective Praise} lesson.  The demographic breakdown of these tutors was as follows: 52\% were White, 18\% Asian, and 52\% male, with over half being 50 years or older. The objective of \textit{Giving Effective Praise} is to equip tutors with skills to boost student motivation through the delivery of effective praise. We collected 129 responses from the tutors who completed the lesson, and these responses are sorted according to the type of praise (i.e., effort-based praise, outcome-based praise, and person-based praise). Notably, the dataset contained only one instance of person-based praise (\textit{``You are very smart''}), leading to its exclusion from the analysis. Thus, our study mainly focused on the analysis of effort-based and outcome-based praise.

\subsection{Sequence Labeling}
\label{labeling}
We aim to provide explanatory feedback that can highlight components of effort-based and outcome-based praise within the tutor responses. Thus, we decided to use a sequence labeling method. By doing so, we created the annotation guideline as well as specific examples of \textit{Effort} and \textit{Outcome} based on the studies \cite{thomas2023tutor, chine2022scenario, chine2022development}. To carry out the annotation work, we hired two expert educators who first completed the lesson of \textit{Giving Effective Praise} from our platform and then started annotating the praise tags representing attributes associated with \textit{Effort} and \textit{Outcome}, for 129 tutor responses.  

\begin{figure}[h]
\centering
\includegraphics[width=0.25\textwidth]{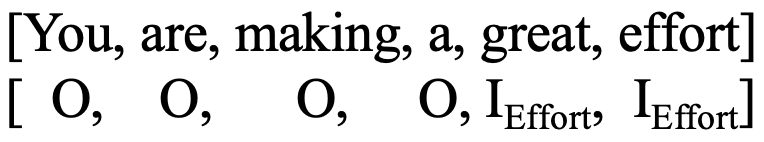}
\caption{Labeling the praise components using IO scheme} \label{token_list}
\end{figure}

In the pursuit of advancing our understanding of effective praise within tutoring dialogues, our study leverages the \textbf{I}nside-\textbf{O}utside (\textbf{IO}) labeling scheme \cite{konkol2015segment} in our study. The \textbf{IO} scheme can capture the necessary information for our analysis, allowing us to maintain focus on the core aspects of praise within tutor responses without the need for differentiating between the beginning or end of entities, which aligns with our needs. The \textbf{IO} scheme, characterized by its simplicity and efficiency, labels tokens as an inside tag (\textbf{I}) and an outside tag (\textbf{O}). The \textbf{I} tag is for the praise components, whereas the \textbf{O} tag is for non-praise words.  For example, when annotating praise components for a tutor's praise \textit{``You are making a great effort''}, the words \textit{``great''} and \textit{``effort''} are identified as part of the \textit{Effort} (i.e., \textbf{I\textsubscript{Effort}}) and the remaining text in the response is identified as the outside (i.e., `\textbf{O}') of the praise components. By annotating the praise components for each tutor response, we can obtain a list of tokens as shown in Figure \ref{token_list}.

In assessing inter-rater reliability for our study, we note that while Cohen’s Kappa is considered the standard measure of inter-annotator agreement for most annotation tasks \cite{mchugh2012interrater}, its suitability for sequence labeling tasks\textemdash Named Entity Recognition or similar tasks where labels are assigned to specific words or tokens within a sequence\textemdash is limited \cite{brandsen2020creating, grouin2011proposal}. Specifically, sequence labeling may result in partial agreements between annotators (e.g., consensus on token label type but not on exact boundaries), which may not be effectively captured by Cohen's Kappa as it does not account for partial agreement \cite{brandsen2020creating}. Additionally, in sequence labeling, a large proportion of tokens are typically labeled as `\textbf{O}' (the distribution of token labels in our study is shown in Table \ref{tab:distribution}), leading to an imbalanced label distribution.  Since Cohen's Kappa assumes an equal likelihood of each category being chosen,  it may not provide a meaningful measure of agreement in situations where the vast majority of labels belong to a single category, making the metric less informative or even misleading \cite{brandsen2020creating}. Given these limitations, F1 score is often preferred for evaluating inter-rater reliability in sequence labeling tasks as suggested in previous studies \cite{brandsen2020creating, deleger2012building}. As the token level Cohen's Kappa scores can also provide some insight, we provide both Cohen’s Kappa and F1 scores to provide a comprehensive view of annotator agreement in our study. Our results\textemdash 0.49 for Cohen's Kappa and 0.79 for the F1 score\textemdash were deemed acceptable for the purposes of our task as suggested by \cite{brandsen2020creating, gisev2013interrater}. To address discrepancies between two annotators, a third expert was invited to resolve inconsistencies. The distribution of annotated praise in our dataset is as follows: 59 responses with only effort-based praise, 22 with only outcome-based praise, 31 containing both types, and 17 lacking mentions of either, illustrating the varied nature of praise within the responses.


\subsection{GPT Facilitated Sequence Labeling}

As discussed, our study employed two widely used approaches for adapting GPT models to sequence labeling tasks: prompt engineering and fine-tuning. Each method offers unique advantages and impacts the process of creating automated explanatory feedback in different ways.

\subsubsection{Prompt engineering for identifying praise components}
\label{prompting_method}
To answer \textbf{RQ} 1, we conducted prompt engineering to design certain prompting strategies to enable GPT models to identify the praise components within the tutor responses. Prompting engineering approach involves designing and structuring input prompts to guide the GPT model in generating desired outputs \cite{liu2023pre, zhou2022large}. The art of prompt engineering lies in crafting prompts that can effectively communicate the context and requirements of the task to the model \cite{liu2023pre, zhou2022large}. In our study, given the presence of tutor responses that exemplify both effort-based and outcome-based praise in the tutor training lesson, we employed a two-shot prompting strategy to guide GPT-3.5 (\texttt{gpt-3.5-turbo-0125}) and GPT-4 (\texttt{gpt-4-0125-preview}) models to highlight praise components within tutor responses. Our prompt is shown in Table \ref{tab:prompt_design}. The following explains our prompt design, aimed at extracting specific elements from tutor responses related to praising student effort and outcomes.

\begin{table}[h]
\centering
\caption{Prompt for identifying praise from tutor responses}
\renewcommand{\arraystretch}{1.3}
\resizebox{0.48\textwidth}{!}{%
\begin{tabular}{|lp{7cm}|}
\hline
\textbf{Role} & \textbf{Content} \\ \hline
\textbf{System} & \textit{You are a response evaluator designed to output JSON. Your task is to analyze tutor responses based on the principles of effective praise focusing on `effort' and `outcome'. Extract words or phrases that represent praise for the student's effort and outcome, and output the results in JSON format with keys titled `Effort' and `Outcome'.} \\ 
\textbf{User} & \textbf{Lesson Principle} \\ 
\textbf{Assistant} & \textit{Sure, can you provide a tutor response for analysis} \\ 
\textbf{User} & \textit{An example of outcome-based praise is: ``Great job! You are a genius!''} \\ 
\textbf{Assistant} & \textit{An output json format is: \{``effort'': [], ``outcome'': [``Great job'']\}} \\ 
\textbf{User} & \textit{Nice, let's do it again.} \\
\textbf{Assistant} & \textit{Sure, can you provide a tutor response for analysis?} \\ 
\textbf{User} & \textit{An example of effort-based praise is: ``You are almost there! I am proud of how you are persevering through and striving to solve the problem. Keep going!''}\\ 
\textbf{Assistant} & \textit{An output json format is: \{``effort'': [``persevering through and striving to solve the problem'', ``Keep going''], ``outcome'': []\}} \\ 
\textbf{User} & \textit{Nice, let's do it again.} \\ 
\textbf{Assistant} & \textit{Sure, can you provide a tutor response for analysis} \\ 
\textbf{User} & \textbf{Tutor Response} \\
\hline
\end{tabular}
}
\label{tab:prompt_design}
\end{table}

\begin{itemize}
    \item \texttt{\{Lesson Principle\}}: This segment provides the guiding principles for desired tutor responses. It includes key aspects of effective praise in educational settings, such as sincerity, specificity, immediacy, authenticity, and focus on the learning process. This principle acts as a reference for evaluating the tutor responses. The lesson principle is detailed in Appendix \ref{learning_principle}
    \item \texttt{\{Tutor Response\}}: This part simulates an interactive environment where the model identify the praise components from the input of tutor responses. 
\end{itemize}



\subsubsection{Fine-tuned GPT models for identifying praise components}

Given limited access to fine-tuning capabilities for the GPT-4 model, we focused on optimizing the use of GPT-3.5 (\texttt{gpt-3.5-turbo-1106}) to answer the \textbf{RQ2}, particularly within the constraints of a modestly sized training dataset. The model fine-tuning approach was implemented to train the GPT-3.5 model to recognize and understand the patterns associated with identifying praise components in tutor responses. To prepare our data for the fine-tuning process, we converted tutor responses and their associated tags into JSON format. This format facilitated the structured representation of our data, mirroring the input style typically expected by the GPT model. The structure of our input data closely resembled the prompts used in GPT model training, with a key distinction: instead of prompting the model to generate text containing praise, we supplied it with annotated outcomes and effort-based praises. Due to the page limit and avoid repetitive content appearing in the paper, we decided to put the details of the fine-tuning input in the Appendix \ref{fine-tune}.

Our approach aimed to investigate the extent to which fine-tuned model can accurately classify and label praise components under limited training dataset, thereby enhancing its performance on our task. By doing so,  we first divided our dataset evenly, allocating 50\% (65 responses) for training and the remaining 50\% (64 responses) for testing. The distribution of annotation is shown in Table \ref{tab:distribution} which presents \textbf{O} as the major tag in our dataset. Then, we subdivided our training set into five distinct partitions: 13, 26, 39, 52, and 65 responses. For each partition, the training process was repeated five times using different random seeds. These partitions represented 10\%, 20\%, 30\%, 40\%, and 50\% of our original dataset, respectively. This stratified approach allowed us to simulate different training conditions, thereby enabling a comprehensive analysis of the model's adaptability and learning efficiency as the amount of available training data varied.

\begin{table}[!htb]
\centering
\caption{Distribution of token labels. }
\label{tab:distribution}
\resizebox{0.45\textwidth}{!}{%
\renewcommand{\arraystretch}{1.3}
\begin{tabular}{lccc}
\toprule
\multirow{2}{*}{}   & \multicolumn{3}{c}{\textbf{\% Annotation (I/O)}}                                              \\ \cmidrule(l){2-4} 
                    & \textbf{O}   &  \textbf{I\textsubscript{Effort}}  & \textbf{I\textsubscript{Outcome}} \\ \midrule
\textbf{Full}       & 2415 (77.72\%)     & 562 (18.09\%)       & 130  (4.18\%)     \\ \midrule
\textbf{Training}   & 1241 (78.99\%)   &   282 (17.95\%)       &   48 (3.06\%)    \\ \midrule
\textbf{Testing}    & 1174 (76.43\%)    &  280 (18.23\%)  &   82 (5.34\%)    \\ \bottomrule
\end{tabular}
}
\vspace{-3mm}
\end{table}

\subsection{Metrics}
\subsubsection{Modified Inter Section over Union Scores}
\label{miou}
In sequence labeling tasks, traditional metrics like the F1 score, as depicted in Equation \ref{eq:f1_score}, are commonly used to assess model performance \cite{brandsen2020creating}. In the context of our study, True Positives (TP) represent the number of tokens correctly identified as praise by the model, False Positives (FP) refer to tokens incorrectly classified as praise, often resulting from the model predicting additional words as part of the praise. False Negatives (FN), on the other hand, are tokens that were part of praise but were overlooked by the model, indicative of missed praise components. Previous research \cite{lin2023using} has highlighted that certain additional entities identified by the model can still contribute meaningfully to human tutors' understanding of response correctness. For instance, as illustrated in Table \ref{tab:different_highlight}, while the first row shows expert annotations of effort-based praise, subsequent examples (rows 2-5) might be model-generated. Notably, rows 2 to 4, despite including additional words for effort-based praise (i.e., FP), offer valuable insights that could assist tutors, contrasting with row 5 where the model's highlighting of merely ``\textit{great}'' (i.e., FN) fails to clearly convey the type of praise intended. This observation suggests the a need for a metric that accommodates the evaluation of additional identified praise tokens more flexibly. However, the F1 score, as shown in Equation \ref{eq:f1_score}, applies the same weight to both FP and FN, a treatment that diverges from our requirement for a more nuanced metric. Consequently, we propose adopting the Intersection over Union (IoU) concept, commonly utilized in the computer vision domain, to better suit our evaluation needs in our task.

\begin{equation}
F1 \; score = \frac{TP}{TP + \frac{1}{2}(FP + FN)} 
\label{eq:f1_score}
\end{equation}

\begin{table}[h]
\centering
\caption{Original training instance and different types of augmented instances. We highlighted outcome-based praise using yellow color and effort-based praise using blue. }
\renewcommand{\arraystretch}{1.5}
\label{tab:different_highlight}
\resizebox{0.43\textwidth}{!}{%
\begin{tabular}{clc}
\hline
    & \textbf{Instance}   & \textbf{Label}                                 \\ \hline
1         & \textit{John, you are making a really \colorbox{effort}{great effort}}. & True \\ \hline
2         & \textit{John, you are making a \colorbox{effort}{really great effort}}. & Pred \\ \hline
3         & \textit{John, you are making \colorbox{effort}{a really great effort}.} & Pred \\ \hline
4         & \textit{John, you are \colorbox{effort}{making a really great effort}.} & Pred \\ \hline
5         & \textit{John, you are making a really \colorbox{effort}{great} effort.} & Pred \\ \hline
\end{tabular}
}
\end{table}

The Intersection over Union (IoU) metric, frequently applied in object detection and segmentation tasks as depicted in Equation \ref{eq:iou_score}, quantifies the extent of overlap between predicted and actual annotations, offering a balanced approach to assess model performance \cite{liu2021span, chen2023boundary}. In the context of sequence labeling, the \textit{`Area of Overlap'} (i.e., TP) corresponds to the tokens the model accurately identifies as praise, whereas the \textit{`Area of Union'} includes all tokens labeled as praise by the model (TP and FP) along with all actual praise tokens in the ground truth (TP and FN). Since we recognize the significance of additionally detected words in our study, we propose a Modified Intersection over Union (M-IOU) metric (shown in Equation \ref{eq:miou_score}) to refine IoU metric further. This modification incorporates a weight coefficient, \(\alpha\), which aims to reduce the influence of FPs on the overall performance score, thus introducing a measure of flexibility towards additional identified words without neglecting the potential for inaccuracies. The coefficient \(\alpha\) is introduced as a real number set at or above zero, enabling users to adjust the tolerance level for additional words identified. A higher \(\alpha\) value enforces a stricter penalty on FPs, while a lower value indicates a more lenient approach. In our analysis, \(\alpha\) is set to 0.2 based on our observation of expert annotations. Notably, in situations where a response lacks praise and the model's prediction concurs (i.e., \(TP + FP + FN\) equals 0), indicating a perfect match between model and ground truth in identifying no relevant praise tokens, we encounter a scenario reflective of novice tutors possibly providing irrelevant responses (e.g., ``\textit{Can I see any of your writing}''). Such irrelevant responses underscore the necessity of our explanatory feedback system in guiding tutors on giving effective praise. For this case,  we adjust the M-IOU formula to directly assign a score of 1 to reflect perfect agreement and underscore the adaptability of our M-IOU in accurately evaluating model precision, particularly in the absence of praise, thus showing its effectiveness in practical applications.

\begin{equation}
\text{IoU} = \frac{\text{Area of Overlap}}{\text{Area of Union}} = \frac{TP}{TP + FP + FN} 
\label{eq:iou_score}
\end{equation}

\begin{equation}
\text{M-IOU} = \frac{TP}{TP + \alpha \times FP + FN}
\label{eq:miou_score}
\end{equation}

\subsubsection{Human annotation and correlation analysis}
\label{human_validation}

To assess the efficacy of our proposed M-IOU score, we undertook a rigorous process involving human annotation to rate the quality of identified components of praise within tutor responses. The human rating scores are further used to compare with our proposed M-IoU score to ensure that M-IoU not only holds computational validity but also aligns with human judgments regarding the praise components in the tutoring responses. Recognizing the importance of human judgment in our study, we hired two additional human coders to scrutinize the highlighted components of praise in tutor responses. These coders attended a detailed annotation training session and completed the lesson of \textit{Giving Effective Praise}, equipping them with the necessary background to perform their evaluations effectively.

Before beginning their rating tasks, we randomized the presentation order of highlighted texts generated by both GPT models and expert annotations for each tutor response. This approach ensured the unpredictability of expert annotation sequence, aiming to mitigate any potential bias in the coders' evaluations. Inspired by the study \cite{shrivastava2020iseql}, we guided the coders to assess each highlighted response based on two questions:  \textbf{Question 1:} \textit{``Do you think the highlighted text provides enough information to identify praise on effort?''} and \textbf{Question 2:} \textit{``Do you think the highlighted text provides enough information to identify praise on the outcome?''}. These questions were designed to capture the coders' assessments of the highlighted texts' adequacy in conveying praise, either for the student's effort or the outcome of their work. The coders were instructed to use a five-point Likert scale for their annotations, with the options being: 1 - \textit{Strongly Disagree}, 2 - \textit{Disagree}, 3 - \textit{Neutral}, 4 - \textit{Agree}, 5 - \textit{Strongly Agree}. 

Upon completing the annotations, we calculated the average score for each response, providing a quantitative measure of the consensus between the coders regarding the effectiveness of the highlighted praise text. To determine the effectiveness of the M-IoU score as a metric for evaluating model predictions, we conducted a correlation analysis using Pearson's r to understand the strength and direction of the linear relationship between the M-IoU scores and the human coders' ratings. The correlation analysis help us understand how well the M-IoU score aligns with human judgment and its potential as a surrogate metric for evaluating the  model performance in identifying praise components.

\begin{figure*}[h]
\centering
\includegraphics[width=0.95\textwidth]{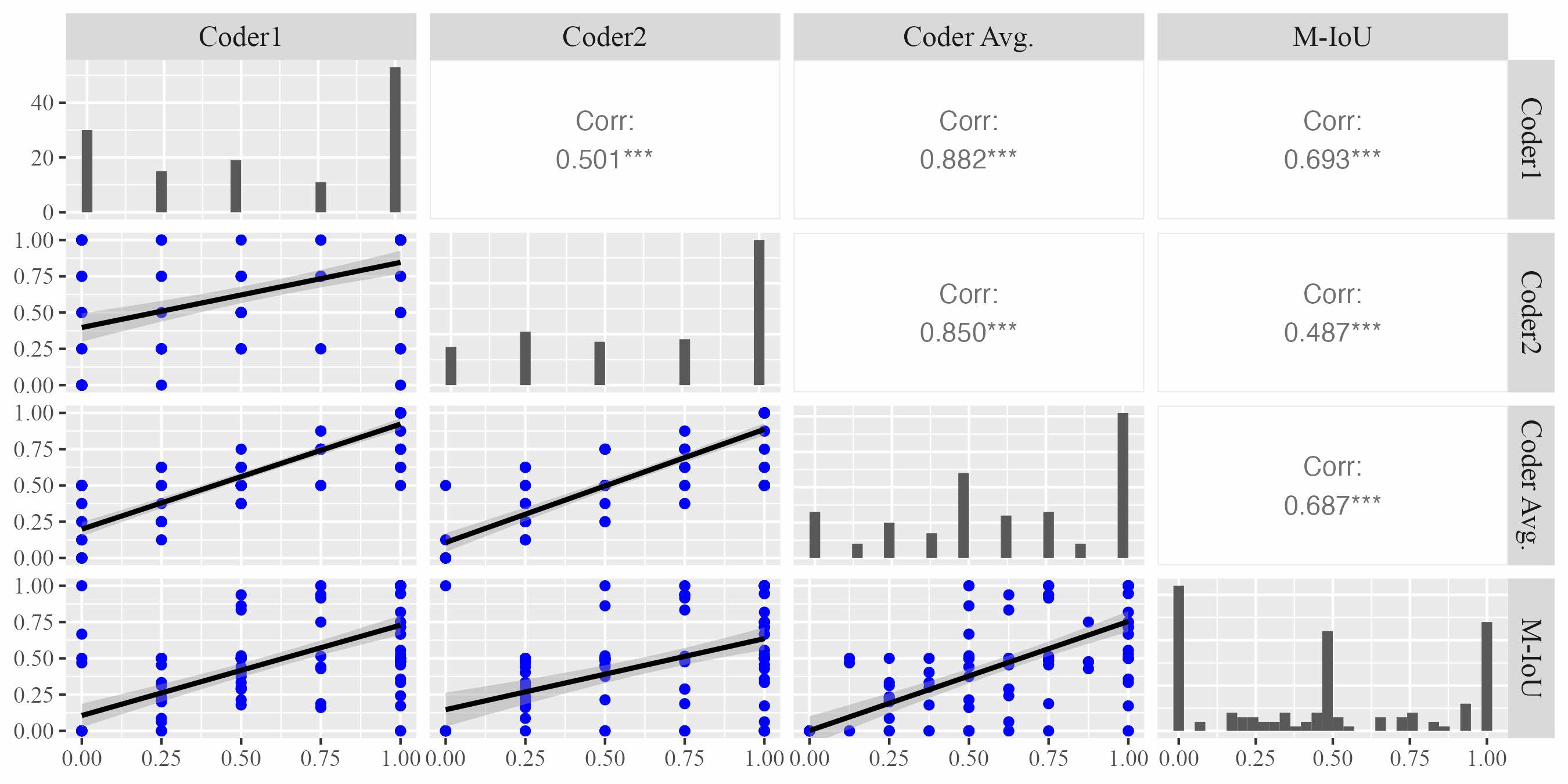}
\caption{Scatter matrix of the correlation on the effort-based praise} \label{result_RQ1_1}
\end{figure*}


\section{Results}
\subsection{Results on RQ1}

To answer \textbf{RQ1}, we aimed to evaluate the extent to which the models' highlighted elements by prompting approach could adequately convey information necessary for identifying the type of praise being expressed. By doing so, we first conducted correlation analysis (as described in the Section \ref{human_validation}) to validate the efficacy of Modified Intersection over Union (M-IoU) score. Due to the page limit,\footnote{The correlation analysis for outcome-based praise can be found in the Appendix \ref{fig:outcome_correlation}.} we only present the results of correlation analysis for effort-based praise between the M-IoU scores and the coders' ratings (shown in Figure \ref{result_RQ1_1}). The findings revealed a significant positive correlation ($p<0.01$) between the M-IoU scores and the ratings from both individual coders (Coder 1 and Coder 2) as well as the averaged scores of the coders (Coder AVG.) for the identification of effort-based praise. This significant correlation underscores the reliability and effectiveness of our proposed M-IoU metric in evaluating the quality of highlighted components from GPT models.

Then, we examined the quality of highlighted elements by prompting GPT-3.5 and GPT-4. In Table \ref{tab:proportion}, we presented the descriptive statistics of the scores rated by two human coders and measured by M-IoU scores. Since the M-IoU score ranging from 0 to 1, to facilitate a direct comparison between human scores and M-IoU scores, we normalized the human coders' rating scores (originally on a scale from 1 to 5) to the same 0 to 1 range. It is important to note that the calculation of M-IoU scores was based on the overlap of highlighted text between the GPT models and expert annotations; consequently, assigning an M-IoU score for expert annotation was not applicable and is thus indicated as N/A in Table \ref{tab:proportion}. The results in Table \ref{tab:proportion} revealed that the highlighted text for outcome-based praise consistently received higher human rating scores and M-IoU scores than that for effort-based praise across both GPT models. This finding aligns with our intuition, considering that outcome-based praise, characterized by expressions such as \textit{``Good job''} and \textit{``Well done,''} tends to be more structured and straightforward to identify than the more nuanced effort-based praise. Interestingly, the difference in M-IoU scores between GPT-3.5 and GPT-4 for both types of praise was marginal, despite the reputed superiority of the GPT-4 model in numerous educational tasks.

\begin{table}[h]
\caption{Descriptive statistics of the scores rated by two human coders and measured by M-IoU scores}
\label{tab:proportion}
\renewcommand{\arraystretch}{1.45}
\resizebox{0.48\textwidth}{!}{%
\begin{tabular}{lcccccc}
\hline
\multirow{2}{*}{\textbf{}} & \multicolumn{2}{c}{\textbf{GPT-3.5 turbo}} & \multicolumn{2}{c}{\textbf{GPT 4 turbo}} & \multicolumn{2}{c}{\textbf{Expert Annotation}} \\ \cline{2-7} 
 & \multicolumn{1}{r}{\textbf{Effort}} & \multicolumn{1}{r}{\textbf{Outcome}} & \multicolumn{1}{r}{\textbf{Effort}} & \multicolumn{1}{r}{\textbf{Outcome}} & \multicolumn{1}{r}{\textbf{Effort}} & \multicolumn{1}{r}{\textbf{Outcome}} \\ \hline
 \multicolumn{7}{l}{\textbf{Comparison between the normalized human ratings and M-IoU scores}}
\\ \hline
\textbf{Coder 1} & ${0.68}_{0.38}$    & ${0.79}_{0.40}$     & ${0.63}_{0.36}$    &   ${0.75}_{0.39}$    & ${0.77}_{0.35}$    & ${0.89}_{0.30}$    \\
\textbf{Coder 2} & ${0.60}_{0.43}$   & ${0.76}_{0.40}$     & ${0.57}_{0.40}$   & ${0.74}_{0.40}$    & ${0.77}_{0.35}$    & ${0.84}_{0.35}$    \\
\textbf{Avg.} & ${0.64}_{0.35}$    & ${0.77}_{0.37}$     & ${0.60}_{0.33}$   &    ${0.75}_{0.38}$   & ${0.77}_{0.29}$   & ${0.87}_{0.29}$   \\ 
\textbf{M-IoU} & ${0.46}_{0.36}$     &  ${0.68}_{0.44}$     & ${0.47 }_{0.38}$    & ${0.64}_{0.46}$    & N/A     & N/A      \\  \hline
 \multicolumn{7}{l}{\textbf{Proportion of human rating `\textit{Agree}' or higher on our scale*}}
\\ \hline
\textbf{Coder 1} & 64.06\% & 76.56\% & 53.13\% & 75.00\% & 73.44\% & 89.06\% \\
\textbf{Coder2} & 53.13\% & 73.44\% & 46.88\% & 71.88\% & 75.00\% & 84.38\% \\
\textbf{Avg.} & 56.25\% & 73.44\% & 53.13\% & 73.44\% & 75.00\% & 85.94\% \\ \hline
\end{tabular}
}
\begin{tabular}[c]{@{}p{25.5em}@{}}{\scriptsize \textit{*Note:} Proportion of rating greater than or equal to `\textit{Agree}' (i.e., agree with the highlight text provides enough information to identify the praise on effort or outcome)}\end{tabular}
\end{table}

We further investigated the proportion of highlights that achieved a rating of 4 or above (corresponding to \textit{`Agree'} or higher on our scale), termed here as `satisfied highlighted text'. The proportion serves as an indicator of the highlights' utility in facilitating the identification of the accurate type of praise. In Table \ref{tab:proportion}, our analysis disclosed that over 50\% of the effort-based praise highlights generated by prompting the GPT-3.5 model were deemed effective by the coders in identifying effort-based praise, whereas for outcome-based praise, the proportion exceeded 70\%. Interestingly, the satisfaction proportion for GPT-3.5's highlights surpassed those of GPT-4, suggesting a nuanced difference in their performance. Moreover, expert annotations were observed to yield the highest satisfaction rates, with over 70\% for effort-based praise and 80\% for outcome-based praise highlights considered satisfactory by the coders. The coders' ratings for expert-annotated text were generally higher, reflecting the expert annotations' authenticity and precision in capturing the essential elements of praise, which indicates the potential limitations of relying solely on Cohen's Kappa for evaluating agreement in sequence labeling tasks. The coders' perceptions affirm the significance of the highlighted texts' quality over mere statistical agreement, indicating that expert annotations, despite a lower Cohen's Kappa, effectively convey the essential attributes of praise within the tutor responses.

\subsection{Results on RQ2}
Building upon the insights gained from the results on \textbf{RQ1}, where the M-IoU was established as a viable proxy for assessing the quality of text highlighted by GPT models, we delved into the potential of fine-tuning the GPT-3.5 model to enhance its performance in identifying praise within tutor responses. Notably, our ability to fine-tune the GPT-4 model was constrained due to access limitations. Consequently, our efforts were concentrated on the GPT-3.5 model, the performance of which is depicted in Figure \ref{result_RQ2_1} and the detailed results of model performance was shown in the Appendix \ref{detailed_fine-tuned}

\begin{figure}[h]
\centering
\includegraphics[width=0.42\textwidth]{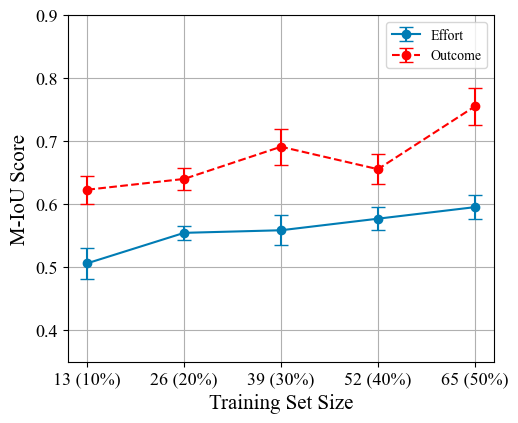}
\caption{Performance of fine-tuning GPT-3.5 model on highlighting correct types of praise in different training size} \label{result_RQ2_1}
\end{figure}

In Figure \ref{result_RQ2_1}, we present the model's performance, quantified by averaging the M-IoU scores derived from five distinct random seeds across various training partitions (from 13 to 65 training sample size). The inclusion of error bars in our analysis offers a visual representation of the model's performance variability, ranging from its maximum to its minimum across these partitions. We simulated a low-resource scenario—characterized by a limited training dataset—and observe the fine-tuned GPT-3.5 model's capability to maintain satisfactory performance under such constraints. Starting with a mere 13 training samples (10\% of the full dataset), the model demonstrated an approximate M-IoU score of 0.5 for effort-based praise and 0.65 for outcome-based praise, showcasing performance on par with that achieved through the prompting method applied to the GPT models. As the training sample size increased, we generally observed an improvement in model performance, with a peculiar exception observed in the outcome-based praise performance when utilizing 52 training samples. Expanding the dataset to 65 samples resulted in the model attaining an M-IoU score of roughly 0.6 for effort-based praise—surpassing the efficacy of the prompting method. Correspondingly, the performance on outcome-based praise reached an M-IoU score of 0.75, rivaling that of expert annotations given that an M-IoU of 0.68 equates to a human rating score of 0.77.

Motivated by these promising outcomes, we elected to adopt the model exhibiting optimal performance in highlighting effort-based praise as the foundation for our automated feedback system within tutor training programs. This decision is underpinned by the pivotal role of effort-based praise in educational feedback; it is the essence of effective praise, determining the appropriateness and impact of the tutor's feedback on student motivation and learning. The ability to accurately identify and underscore effort-based praise in tutors' responses is thus deemed crucial for enhancing the quality of educational feedback. In support of this initiative, a demo of our automated explanatory feedback system is accessible via the provided link\footnote{The demo of our automated explanatory feedback can be found here \url{https://edm24-effort-outcome.vercel.app/}}, showcasing the application's potential to transform tutor training by emphasizing the significance of effort-based praise.

\section{Discussion}
Our study examined the potential of GPT models to highlight the desired and undesired parts of praise from trainee responses and further integrated the highlighted parts into the feedback for tutor training. By employing the Modified Intersection over Union (M-IoU) as a novel metric, we measured the quality of the praise highlighted by GPT models. The M-IoU metric, validated through correlation with human coders, underscores the potential of combining human intuition with algorithmic metrics to enhance the specificity and relevance of educational feedback. The findings from our investigation confirmed the considerable promise of employing techniques such as \textit{prompting} and \textit{fine-tuning} within GPT models to generate automated, explanatory feedback tailored for tutor training programs. By leveraging a fine-tuned GPT model, we developed an automated feedback system specifically designed for tutor training, with the objective of delivering immediate and explanatory feedback. This innovation presented a viable, scalable solution to the pressing challenge of delivering personalized feedback to learners (trainee tutors as learners in our study). The implementation of an automated explanatory feedback system in our study exemplifies how such technology can be leveraged to identify specific elements in tutors' open-ended responses that are either desirable or in need of enhancement.

\textbf{Prompting GPT model to highlight key components.} Upon evaluating the highlighted praise components from GPT models (prompting) and expert annotations, we observe that the quality of highlighted praise components by experts typically outperform those highlighted by GPT models. For instance, as indicated in the first row of Table \ref{tab:examples}, there is unanimous agreement among human coders that the highlighted praise components by expert annotations are better than those highlighted by both GPT models. Specifically, while the GPT-3.5 model accurately identified phrases such as \textit{``doing a great job''} and \textit{``Stick with this''} as forms of praise, it erroneously categorized \textit{``doing a great job''} as effort-based praise, contrary to the established praise principle which classifies it as outcome-based praise \cite{thomas2023tutor}. Conversely, the GPT-4 model correctly classified \textit{``doing a great job''} as outcome-based praise but included additional words like \textit{``We can finish it''} in its identification of effort-based praise. The comparison of additional words included in effort-based praise annotations between GPT-3.5 and GPT-4 models resulted in identical scores from the coders (0.5 in Table \ref{tab:examples}, reflecting a neutral stance equivalent to a score of 3 on the Likert Scale). The identical scoring stems from the equal number of additional words identified by both models. Furthermore, the M-IoU score aligns with the coders' assessments, underscoring the metric's utility in capturing the accuracy of the models' annotations. In another observation, detailed in the second row of Table \ref{tab:examples}, both coders concurred that in certain instances, prompting GPT-3.5's identification of praise components was superior to that of expert annotations. Additionally, the third row of Table \ref{tab:examples} presents a scenario with significant discrepancies in the ratings assigned to the highlighted praise components by GPT-3.5 and GPT-4 between two coders. Here, the M-IoU score proved instrumental in mitigating the variances in individual assessments, effectively approximating the average score derived from both coders' ratings.

\begin{table*}[htb]
\centering
    \caption{Examples of evaluation on highlighted praise components from two human coders' rating scores and M-IoU scores.}
\label{tab:examples}
\renewcommand{\arraystretch}{1.3}
\resizebox{0.95\textwidth}{!}{%
\begin{tabular}{@{}clp{10cm}cccccc@{}}
\toprule
\multirow{2}{*}{\textbf{Row}} & \multirow{2}{*}{\textbf{Categories}} & \multirow{2}{*}{\textbf{Responsese}} & \multicolumn{2}{c}{\textbf{Coder 1}} & \multicolumn{2}{c}{\textbf{Coder 2}} & \multicolumn{2}{c}{\textbf{M-IoU}} \\ \cmidrule(l){4-9} 
 &  &  & \textbf{Effort} & \textbf{Outcome} & \textbf{Effort} & \textbf{Outcome} & \textbf{Effort} & \textbf{Outcome} \\ \midrule
\multirow{3}{*}{1} & GPT-3.5 & \textit{Carla you are \colorbox{effort}{doing a great job}! \colorbox{effort}{Stick with this}. We can finish it}. & 0.50 & 0 & 0.50 & 0 & 0.50 & 0\\
 & GPT-4 & \textit{Carla you are \colorbox{outcome}{doing a great job}! \colorbox{effort}{Stick with this}. \colorbox{effort}{We can finish it}.} & 0.50 & 0.75 & 0.50 & 0.75  & 0.50 & 0.83 \\
 & Expert & \textit{Carla you are doing a \colorbox{outcome}{great job}! \colorbox{effort}{Stick with this}. We can finish it.} & 1.00 & 1.00 & 1.00 & 1.00  & N/A & N/A \\ \midrule
\multirow{3}{*}{2} & GPT-3.5 & \textit{\colorbox{outcome}{Great job}, Kevin! I can tell \colorbox{effort}{how hard you worked to get there}.} & 1.00 & 1.00 & 1.00 & 1.00  & 0.53 & 1.00 \\
 & GPT-4 & \textit{\colorbox{outcome}{Great job}, Kevin! \colorbox{effort}{I can tell how hard you worked} to get there.} & 0.75 & 1.00 & 0.75 & 1.00  & 1.00 & 1.00 \\
 & Expert & \textit{\colorbox{outcome}{Great job}, Kevin! \colorbox{effort}{I can tell how hard you worked} to get there.} & 0.75 & 1.00 & 0.75 & 1.00  & N/A & N/A \\ \midrule
\multirow{5}{*}{3} & GPT-3.5 & \textit{\colorbox{outcome}{Great job} Kevin! \colorbox{effort}{Your determination is really admirable! Pretty sure} \colorbox{effort}{you can complete it with this determination}!} & 0.25 & 1.00 & 1.00 & 1.00  & 0.48 & 1.00 \\
 & GPT-4 & \textit{\colorbox{outcome}{Great job} Kevin! \colorbox{effort}{Your determination is really admirable! Pretty sure} \colorbox{effort}{you can complete it with this determination}!} & 0.25 & 1.00 & 1.00 & 1.00  & 0.48 & 1.00 \\
 & Expert & \textit{\colorbox{outcome}{Great job} Kevin! Your \colorbox{effort}{determination is really admirable}! Pretty sure you can complete it with this determination!} & 1.00 & 1.00 & 0.25 & 1.00  & N/A & N/A \\ \bottomrule
\end{tabular}
}
\end{table*}

\textbf{Fine-tuning GPT model to highlight key components.} Then, our study assessed the impact of fine-tuning the GPT-3.5 model with different amount of training data to determine the optimal dataset size required to achieve satisfactory performance in generating explanatory feedback. This insight is important for researchers and educational practitioners seeking to use LLMs effectively, especially when faced with constraints on data availability. Our findings highlight the critical role of task-specific optimization for LLMs, illustrating how strategic modifications to the quantity of training data can markedly enhance the performance of automated feedback systems. By identifying the minimum dataset requirements for fine-tuning GPT models, our study provides valuable guidelines for developing effective explanatory feedback.
Furthermore, by integrating our proposed prompting strategies alongside a certain number of training datasets, we found that the fine-tuned GPT-3.5 model generally outperforming prompting models (both GPT-3.5 and GPT-4) in identifying the praise elements (including effort- and outcome-based praise). It suggest that, despite the general advancements represented by newer models like GPT-4, fine-tuning earlier versions such as GPT-3.5 can achieve comparable or even superior performance in specific applications. This insight is important for educational practitioners and researchers, particularly those constrained by financial limitations, as fine-tuning GPT-3.5 proves to be a more cost-effective option than prompting GPT-4. Moreover, the fine-tuning approach offers a solution to challenges related to accessing the latest models or dealing with limited resources, such as a restricted number of training datasets.

\textcolor{royal_purple}{\textbf{Comparison of prompting and fine-tuning approaches.} 
In our study, we employed both prompting and fine-tuning approaches to adapt large language models, specifically GPT models, for providing highlighting the desired and undesired parts of trainee responses. Prompting enables rapid model adaptation to highlight the components of effort- and outcome-based praise without extensive retraining, thus conserving computational resources and time. However, since the model parameters are not updated, prompting might not capture deeper insights from annotated data, potentially limiting performance on highlighting key components from complex responses. For example, consider the tutor response ``\textit{Great job figuring out that problem! Would you like help with anything else?}'' Because of the use of ``\textit{figuring out}'', this response is categorized as effort-based praise, however, GPT-4 without fine-tuning mistakenly classified it as outcome-based, whereas GPT-3.5 with fine-tuning correctly classified it as effort-based. This error likely occurred because the model over-weighted the generic phrase ``\textit{Great job}''. Additionally, while the prompting approach offers flexibility in testing different prompts to quickly gauge the model's capabilities on our task, its effectiveness heavily depends on the quality of the prompt design. As observed during our prompt engineering phase, inadequate prompts can lead to misleading outputs.}

\textcolor{royal_purple}{On the other hand, fine-tuning allows for deeper model customization by adjusting internal parameters to closely align with our task in identifying the components of praises from tutor responses, often resulting in superior performance measured by M-IOU scores, as observed in our study. Fine-tuning enables the GPT model to deeply integrate new knowledge and adjust its existing knowledge, better fitting the task requirements of identifying components of effort- and outcome-based praise. Despite these advantages, fine-tuning requires a substantial amount of relevant and high-quality data and significant computational resources. The data must be carefully annotated to guide the model effectively toward the desired behavior, which present a significant limitation if such data is scarce or difficult to collect. Additionally, fine-tuning involves updating the weights of a neural network based on a specific dataset, a process that can be resource-intensive and requires access to powerful hardware, especially for larger models.}

\textcolor{royal_purple}{To address some of these challenges and further enhance our highlighted feedback system, we are considering the integration of Retrieval-Augmented Generation (RAG). RAG combines the strengths of both retrieval and generation models to improve the performance of language models on specific tasks \cite{lewis2020retrieval}. RAG could enhance the performance of prompting LLMs by dynamically incorporating relevant external information into responses, providing more informed and contextually accurate outputs (e.g., \cite{han2024improving}). Additionally, RAG can be integrated with the fine-tuning approach for providing highlighted feedback, potentially improving the model's accuracy in highlighting components of praise. This integration aims to create a model that not only leverages external data through RAG but also adapts more finely to specialized tasks through fine-tuning, demonstrating superior performance in contextually rich and dynamic environments.}

\section{Limitation and Future Works}

\textcolor{royal_purple}{\textbf{Measuring the impact of the proposed feedback system.} While the current study demonstrates the potential of using GPT-based models for providing explanatory feedback in a novice tutor training context, we acknowledge the necessity of validating the effectiveness of feedback with highlighted components through empirical research involving actual users. To this end, we propose a comprehensive study aimed at assessing the real-world effectiveness and impact of our feedback system on novice tutors. The planned study will involve a group of novice tutors who will use our automated feedback system during their training sessions. The study will be designed to capture both qualitative and quantitative data to provide a holistic evaluation of the feedback system's performance. Quantitative data will be collected through pre-and post-tests to measure the learning gains of tutors, while qualitative data will be gathered from surveys and interviews to assess tutors' perceptions and experiences with the feedback.}

\textcolor{royal_purple}{\textbf{Expanding the scope of the proposed feedback systems for diverse tutoring scenarios.}} We aim to empower novice tutors through automated explanatory feedback, enabling them to grasp effective tutoring strategies within our training programs. While the fine-tuned GPT-3.5 model has shown promising results in delivering explanatory feedback for giving effective praise, its applicability and effectiveness across a broader range of tutoring scenarios, such as responding to student errors and assessing student understanding, have yet to be explored. This gap highlights the necessity of broadening the scope of our proposed method. Expanding and rigorously evaluating our approach to encompass diverse educational contexts and lesson types is essential for building a more versatile and universally applicable automated feedback system. 

\textcolor{royal_purple}{\textbf{Enhancing the proposed feedback system with data augmentation.}} We also recognized the inherent challenges associated with sequence labeling for highlighting key components of tutoring practice (e.g., praise components in our study). To achieve satisfactory performance, our study required the use of 50\% of the total dataset, equivalent to 65 training samples. This substantial annotation workload raises concerns, particularly when considering the extension of fine-tuning GPT models to more tutor training lessons (e.g, our tutor training platform has designed 20 lessons for different tutoring strategies). To address this issue and reduce the reliance on extensive manual annotation, we are exploring the implementation of data augmentation techniques, such as random swap and synonym replacement \cite{feng2021survey}. By applying these data augmentation techniques to merely 10\% of the dataset or 13 training samples, we aim can reduce the dependency on extensive manual annotation efforts.


\textcolor{royal_purple}{\textbf{Examining the applicability of the proposed feedback system across different platforms.}} In our future work, we aim to apply sequence labeling methods to analyze real-world tutoring transcripts and diverse datasets, such as teacher comments from educational platforms like ASSISTments \cite{heffernan2014assistments}. Leveraging fine-tuned GPT models on highlighting the key components of instructional strategies (e.g., effective praise, response to student errors, and engaging with difficult students), we plan to generate comprehensive reports that highlight the desired and less desired components from the teacher feedback or comments and provide targeted feedback with suggestions for improvements. This initiative will potentially offer actionable insights to tutors on enhancing their pedagogical approaches in future sessions.



\section{Conclusion}
In this study, we investigated the enhancement of automated feedback systems through the application of GPT models, employing a multifaceted approach that included the utilization of prompting GPT-3.5 and GPT-4 models and fine-tuning GPT-3.5 models for improved performance. Prompting GPT models demonstrated their potential in guiding models to identify specific components of praise, emphasizing the critical role of prompt design in optimizing model outputs. In comparison, fine-tuning the GPT-3.5 model, in particular, significantly enhanced the system's ability to accurately highlight key components from tutor responses. This led to the development of an automated feedback system aimed at delivering immediate and explanatory feedback for tutor training, addressing the crucial need for scalable and effective feedback. Our implementation showcases the potential of leveraging advanced large language models to provide highlighting explanatory feedback on tutors' open-ended responses, offering insights for future research in the development of automated feedback systems.


\section{Acknowledgments}
\textcolor{royal_purple}{This work is supported by funding from the Richard King Mellon Foundation (Grant \#10851) and the Learning Engineering Virtual Institute (\href{https://learning-engineering-virtual-institute.org/}{https://learning-engineering-virtu\\al-institute.org/}). Any opinions, findings, and conclusions expressed in this paper are those of the authors. We also wish to express our gratitude to Dr. Ralph Abboud and Dr. Carolyn P. Ros{\'e} for their invaluable guidance and recommendations, and to Yiyang Zhao and Yuting Wang for their assistance in verifying the rating scheme.}

%
\bibliographystyle{abbrv}
\bibliography{sigproc}  
%

\clearpage
\newpage
\appendix
\section{Lesson Principles}
\label{learning_principle}
The following is the principle that a correct response should follow:\\
Praising students for working hard and putting forth effort is a great way to increase student motivation. When the learning gets tough, giving correct praise is a powerful strategy to encourage students to keep going.\\
The correct response should be : \\
-perceived as sincere, earned, and truthful.\\
-specific by giving details of what the student did well.\\
-immediate with praise given right after the student action.\\
-authentic and is not repeated often, such as “great job” which loses meaning and becomes predictable.\\
-focused on the learning process, not ability
(AJTutoring.com, 2022)\\
Correct responses must follow some, but not all the above. \\
There are two types of praise responses: Effort and Outcome praise \\
- Effort praise focuses on the learning process. Effort praise recognizes students for putting forth effort and persevering through the learning process instead of focusing on whether a student got the problem correct or pure ability.\\
- Outcome praise showcases student's achievements, such as getting a grade A on an assignment or getting a problem correct, and is often, but not always, associated with unproductive praise.\\
To receive full credit of correct praise, tutors cannot just say "great job" and praise with no specific reasoning. Tutors need to praise for effort AND be positive and encouraging.\\

\newpage
\section{Input for Fine-tuning GPT-3.5}
\label{fine-tune}

\begin{table}[h]
\centering
\caption{Input for fine-tuning GPT 3.5 model}
\renewcommand{\arraystretch}{1.3}
\resizebox{0.48\textwidth}{!}{%
\begin{tabular}{|lp{7cm}|}
\hline
\textbf{Role} & \textbf{Content} \\ \hline
\textbf{System} & \textit{You are a response evaluator designed to output JSON. Your task is to analyze tutor responses based on the principles of effective praise focusing on `effort' and `outcome'. Extract words or phrases that represent praise for the student's effort and outcome, and output the results in JSON format with keys titled `Effort' and `Outcome'.} \\ 
\textbf{User} & \textbf{Lesson Principle} \\ 
\textbf{Assistant} & \textit{Sure, can you provide a tutor response for analysis} \\ 
\textbf{User} & \textit{An example of outcome-based praise is: ``Great job! You are a genius!''} \\ 
\textbf{Assistant} & \textit{An output json format is: \{``effort'': [], ``outcome'': [``Great job'']\}} \\ 
\textbf{User} & \textit{Nice, let's do it again.} \\
\textbf{Assistant} & \textit{Sure, can you provide a tutor response for analysis?} \\ 
\textbf{User} & \textit{An example of effort-based praise is: ``You are almost there! I am proud of how you are persevering through and striving to solve the problem. Keep going!''}\\ 
\textbf{Assistant} & \textit{An output json format is: \{``effort'': [``persevering through and striving to solve the problem'', ``Keep going''], ``outcome'': []\}} \\ 
\textbf{User} & \textit{Nice, let's do it again.} \\ 
\textbf{Assistant} & \textit{Sure, can you provide a tutor response for analysis} \\ 
\textbf{User} & \textbf{Tutor Response} \\
\textbf{Assistant} & \textbf{Praise Type and Content} \\ 
\hline
\end{tabular}
\label{tab:prompt_design_finetuning}
}
\begin{tabular}[c]{@{}p{25.5em}l@{}}{\small \textit{Note:} \textbf{Praise Type and Content}: This part simulates an interactive environment where the model plays the role of a response evaluator. The conversation flow is designed to mimic a real-world interaction, with system and user roles alternately providing context, instruction, and input (the tutor response) for processing.}\end{tabular}
\end{table}


\clearpage{}

\begin{figure*}[h!]
\section{Scatter matrix of the correlation on the outcome-based praise} \label{fig:outcome_correlation}
\includegraphics[width=0.98\textwidth]{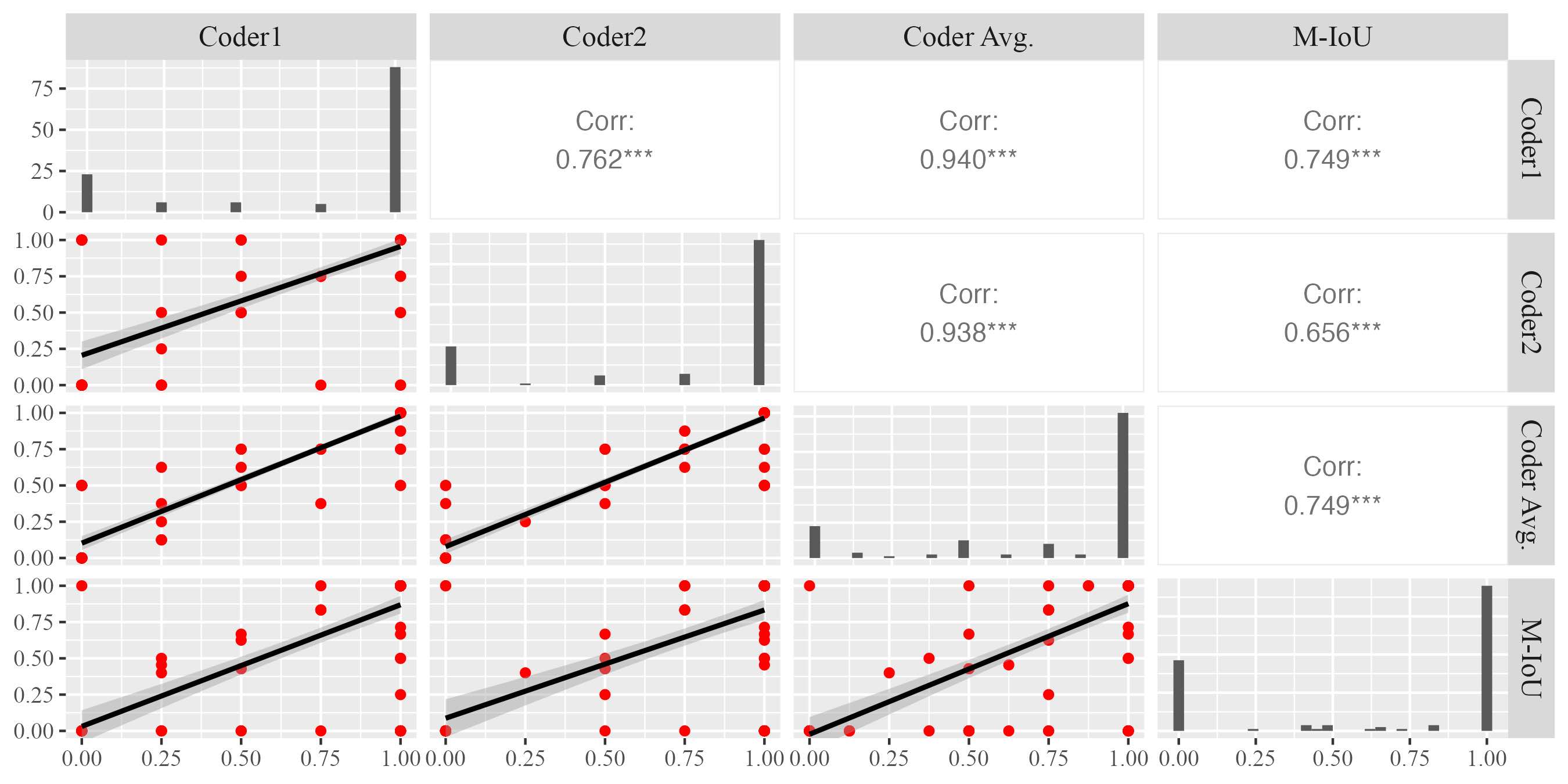}
\caption{Scatter matrix of the correlation on the outcome-based praise} \label{outcome_correlation}
\end{figure*}

\begin{table*}[]
\section{Detailed results of fine-tuned GPT-3.5 models' performance} \label{detailed_fine-tuned}
\centering
\caption{Detailed results of fine-tuned GPT-3.5 models on identifying praise components}
\renewcommand{\arraystretch}{1.3}
\label{tab:detailed_fine-tuned}
\begin{tabular}{lcccccccc}
\hline
\multirow{2}{*}{\textbf{Training size}} & \multicolumn{4}{c}{\textbf{Effort}} & \multicolumn{4}{c}{\textbf{Outcome}} \\ \cline{2-9} 
 & \textbf{Mean} & \textbf{Std.} & \textbf{Min.} & \textbf{Max.} & \textbf{Mean} & \textbf{Std.} & \textbf{Min.} & \textbf{Max.} \\ \hline
\textbf{13 (10\%)} & 0.51 & 0.05 & 0.44 & 0.58 & 0.62 & 0.05 & 0.56 & 0.69 \\ \hline
\textbf{26 (20\%)} & 0.55 & 0.03 & 0.51 & 0.58 & 0.64 & 0.04 & 0.59 & 0.68 \\ \hline
\textbf{39 (30\%)} & 0.56 & 0.05 & 0.46 & 0.59 & 0.69 & 0.06 & 0.59 & 0.75 \\ \hline
\textbf{52 (40\%)} & 0.58 & 0.04 & 0.54 & 0.63 & 0.66 & 0.05 & 0.60 & 0.73 \\ \hline
\textbf{65 (50\%)} & 0.60 & 0.04 & 0.54 & 0.64 & 0.76 & 0.07 & 0.66 & 0.84 \\ \hline
\end{tabular}
\end{table*}

\end{document}